\pdfoutput=1

\documentclass[11pt]{article}

\usepackage{ACL2023}

\usepackage{times}
\usepackage{latexsym}

\usepackage[T1]{fontenc}

\usepackage[utf8]{inputenc}

\usepackage{microtype}

\usepackage{inconsolata}

\usepackage{graphicx}
\usepackage{epstopdf}
\usepackage{hyperref}
\usepackage{xstring}
\usepackage{color}
\usepackage{amsmath}
\usepackage{amssymb}
\usepackage{multirow}
\usepackage{makecell}
\usepackage{CJKutf8}
\usepackage{subcaption}
\usepackage[mathscr]{euscript}
\usepackage{arydshln}

\DeclareMathOperator*{\argmax}{argmax}

\def\tabref#1{Table~\ref{#1}}
\def\figref#1{Figure~\ref{#1}}

\def\secref#1{Section~\ref{#1}}
\def\revise#1{#1} 

\title{Towards Flow Graph Prediction of Open-Domain Procedural Texts}

\author{
      Keisuke Shirai$^1$ \quad
      Hirotaka Kameko$^2$ \quad 
      Shinsuke Mori$^2$ \\
      $^1$Graduate School of Informatics, Kyoto University \\ 
      $^2$Academic Center for Computing and Media Studies, Kyoto University \\
      \texttt{shirai.keisuke.64x@st.kyoto-u.ac.jp} \quad 
      \texttt{\{kameko,forest\}@i.kyoto-u.ac.jp} \\
}

\begin{document}
\maketitle

\begin{abstract}
Machine comprehension of procedural texts is essential for reasoning about the steps and automating the procedures. However, this requires identifying entities within a text and resolving the relationships between the entities. Previous work focused on the cooking domain and proposed a framework to convert a recipe text into a flow graph (FG) representation. In this work, we propose a framework based on the recipe FG for flow graph prediction of open-domain procedural texts. To investigate flow graph prediction performance in non-cooking domains, we introduce the wikiHow-FG corpus from articles on wikiHow, a website of how-to instruction articles. In experiments, we consider using the existing recipe corpus and performing domain adaptation from the cooking to the target domain. Experimental results show that the domain adaptation models achieve higher performance than those trained only on the cooking or target domain data.
\end{abstract}

\section{Introduction}
A procedural text guides a human to complete daily activities like cooking and furniture assembly. Machine comprehension of these texts is essential for reasoning about the steps~\citep{zhang2020reasoning} and automating the procedures~\citep{bollini2013interpreting}. However, it needs to identify entities within a text and resolve relationships between the entities. Converting the text into an actionable representation (e.g., flow graph~\citep{momouchi1980control}) is an approach for solving these problems.

There are several works on converting a procedural text into an action graph~\citep{mori2014flow,kulkarni2018annotated,kuniyoshi2020annotating}. In the cooking domain, various approaches~\citep{mori2014flow,kiddon2015mise,pan2020multimodal,papadopoulos2022learning} have been taken because there are a rich amount of available resources on the web. Among them, recipe flow graph (FG)~\citep{mori2014flow} has the advantage of capturing fine-grained relationships at entity-level. While the original work~\citep{mori2014flow} introduced a framework and corpus in Japanese, the recent work~\citep{yamakata2020english} proposed those in English. The current FG framework has two issues. Since FG is designed to represent the flow of actions in a procedural text, it should be applicable to other procedural domains such as crafting. One is that the framework has only been applied to the cooking domain. The other is that preparing a large number of annotations (e.g., thousands of articles) is unrealistic due to its complex annotation procedures.

\begin{figure}[t]
    \centering
    \includegraphics[scale=0.46]{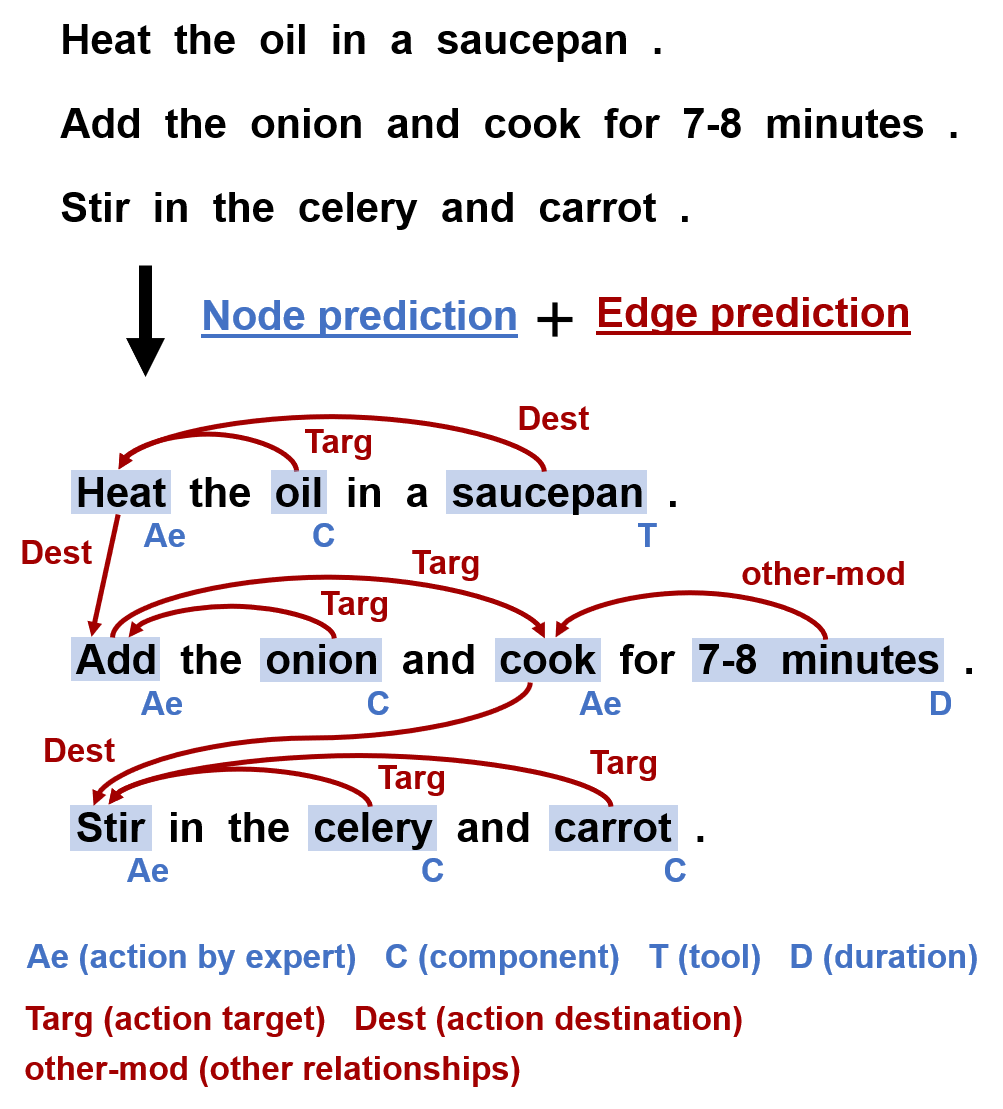}
    \caption{Example of flow graph prediction with our framework. The prediction is in two stages: node prediction (colored in blue) and edge prediction (colored in red). In this work, we use this framework to predict flow graphs of open-domain procedural texts.}
    \label{fig:flow_graph}
\end{figure}
In this work, we propose a framework based on the English recipe flow graph (English-FG)~\citep{yamakata2020english} for FG prediction of open-domain procedural texts. We show the overview of our framework in \figref{fig:flow_graph}. Our motivation is to expand the scope of the recipe FG to non-cooking domains by treating food ingredients as final product components. Our framework predicts an FG in two stages: node prediction and edge prediction, following \citet{maeta2015framework}. Our framework is compatible with the English-FG, and we can jointly learn representations from data in the cooking and non-cooking domains. To investigate FG prediction performance in non-cooking domains, we introduce the wikiHow-FG corpus from wikiHow articles. The corpus was constructed by selecting four domains from the wikiHow categories and annotating $30$ articles for each domain.

\begin{table}[t]
    \begin{center}
        \begin{tabular}{l|l} \hline
            Tag & Meaning \\\hline
            \textsf{C} (\textsf{F}) & Component (Food) \\
            \textsf{T} & Tool  \\
            \textsf{D} & Duration  \\
            \textsf{Q} & Quantity \\
            \textsf{Ae} (\textsf{Ac}) & Action by expert (chef) \\
            \textsf{Ae2} (\textsf{Ac2}) & Discontinuous Ae (Ac) \\
            \textsf{Ac} (\textsf{Af}) & Action by component (food) \\
            \textsf{At} & Action by tool \\
            \textsf{Sc} (\textsf{Sf}) & State of component (food) \\
            \textsf{St} & State of tool \\\hline
        \end{tabular}
    \end{center}
    \caption{Tags and their meanings. The inside of the parenthesis represents a tag and its meaning in English-FG.}
    \label{tab:tag_types}
\end{table}
In experiments, we assume a low-resource scenario in which we can access only a few training examples in the target domain. This is a realistic scenario considering the huge annotation cost for FG. To tackle this issue, we consider domain adaptation from the existing cooking domain to the target domain. Experimental results show that domain adaptation models obtain higher performance than those trained only on the cooking or target domain data. We also considered two data augmentation techniques to boost performance. From the results, we found that they improve performance in particular domains.

Our contributions are three-fold:
\begin{itemize}
    \item We propose a framework based on the English-FG for flow graph prediction of open-domain procedural texts.
    \item We introduce the wikiHow-FG corpus, a new corpus from wikiHow articles. This corpus is based on four wikiHow domains and has $30$ annotated articles for each domain.
    \item We assume a low-resource scenario in the target domain and consider domain adaptation from the cooking to the target domain. Experimental results show that domain adaptation models outperform those trained only on the cooking or target domain data.
\end{itemize}

\begin{table}[t]
    \centering
    \scalebox{0.83}{
        \begin{tabular}{l|l} \hline
            \textsf{Label} & Meaning \\\hline
            \textsf{Agent} & Action agent \\
            \textsf{Targ} & Action target \\
            \textsf{Dest} & Action destination \\
            \textsf{T-comp} & Tool complement \\
            \textsf{C-comp} (\textsf{F-comp}) & Component (Food) complement \\
            \textsf{C-eq} (\textsf{F-eq}) & Component (Food) equality \\
            \textsf{C-part-of} (\textsf{F-part-of}) & Component (Food) part-of \\
            \textsf{C-set} (\textsf{F-set}) & Component (Food) set \\
            \textsf{T-eq} & Tool equality \\
            \textsf{T-part-of} & Tool part-of \\
            \textsf{A-eq} & Action equality \\
            \textsf{V-tm} & Head of clause for timing \\
            \textsf{other-mod} & Other relationships \\\hline
        \end{tabular}
    }
    \caption{Labels and their meanings. The inside of the parenthesis represents a label and its meaning in English-FG.}
    \label{tab:label_types}
\end{table}
\section{Recipe flow graph}\label{sec:recipe_flow_graph}
In this section, we provide a brief description of the recipe flow graph (FG)~\citep{mori2014flow}. A recipe FG is a directed acyclic graph $G(V, E)$, where $V$ represents entities as nodes, while $E$ represents the relationships between the nodes as labeled edges. Currently, FG annotations are available in Japanese~\citep{mori2014flow} and English~\citep{yamakata2020english}, and these corpora provide annotations of hundreds of recipes. 
Note that Japanese and English frameworks for FG are not compatible since the English FG uses additional tags to handle English-specific expressions. As we focus on texts in English, we consider the English-FG framework~\citep{yamakata2020english} in the following sections.

\subsection{Flow graph representation}
A recipe FG representation is divided into two types of annotations; node and edge annotations. Nodes represent entities with tags in the IOB-format~\citep{ramshaw1995text}. As listed in \tabref{tab:tag_types}, $10$ types of tags are used in the English-FG. Labeled edges represent the relationships between the nodes. As listed in \tabref{tab:label_types}, $13$ types of labels are used in the English-FG. 

\subsection{Flow graph prediction}
For the automatic prediction of the FG, previous work~\citep{maeta2015framework} proposed to divide the problem into two subtasks: node prediction and edge prediction. In both subtasks, models are trained in a supervised fashion.

\paragraph{Node prediction} identifies nodes in an article with the tags. \citet{maeta2015framework,yamakata2020english} formulated this problem as a sequence labeling problem and used NER model~\citep{lample2016neural}. While predicting tags at sentence-level is common in NER~\citep{lample2016neural}, previous work~\citep{yamakata2020english} used an entire recipe text as input.\footnote{In our preliminary experiments, we found that predicting the tags at document-level improves accuracy by 10\% compared to the prediction at sentence-level.}

\paragraph{Edge prediction} constructs a directed acyclic graph by predicting labeled edges between the nodes. This is formulated as a problem of finding the maximum spanning tree as:
\begin{equation}
    \hat{G} = \argmax_{G \in \mathscr{G}} \sum_{(u, v, l)} s(u, v, l),
\end{equation}
where $s(u, v, l)$ represents the score of a labeled edge from $u$ to $v$ with label $l$. We can solve this problem by using the Chu-Liu-Edmonds algorithm. The scores are calculated using a graph-based dependency parser~\citep{mcdonald2005non}.

\section{Flow graph prediction of open-domain procedural texts}
Our framework is based on the English-FG and applies to non-cooking domains by treating foods in recipe texts as final product components. Examples of the components include tomato and beef for cooking, cardboard and glue for crafting, and gear and tire for vehicle maintenance. The framework uses tags and labels defined in \tabref{tab:tag_types} and \tabref{tab:label_types}, respectively. These tags and labels are slightly modified from the definitions in the English-FG to avoid confusion, and we did not add or delete any tags and labels. Therefore, our framework is compatible with the English-FG, and we can learn representations jointly from the cooking and non-cooking domains. 

With this framework, we consider predicting flow graphs of open-domain procedural texts. The prediction is performed in two stages: node prediction and edge prediction as in \secref{sec:recipe_flow_graph}. Models are trained in a supervised way as the previous approach~\citep{maeta2015framework}. However, preparing a large number of examples in a new domain is unrealistic, considering the huge FG annotation cost. Thus, we assume that only a few training examples are available in the target domain. To tackle this issue, we consider domain adaptation from the existing cooking domain data to the target domain data. In the rest of this section, we formulate the task in \secref{subsec:task_definition}, then consider data augmentation techniques that fit our setting in \secref{subsec:data_augmentation}.

\subsection{Task definition}\label{subsec:task_definition}
We are given $N$ examples in the cooking domain ${(V^{C}_{1}, E^{C}_{1}), \cdots, (V^{C}_{N}, E^{C}_{N})}$ and $M$ examples in the target domain ${(V^{T}_{1}, E^{T}_{1}), \cdots, (V^{T}_{M}, E^{T}_{M})}$, where $V^{C}$ and $E^{C}$ are a set of vertices and edges in the cooking domain, while $V^{T}$ and $E^{T}$ are those in the target domain. Our goal is to maximize the performance of node and edge prediction models in the target domain. In this work, we use the English-FG corpus (\revise{$300$ articles}) as the cooking domain examples and the wikiHow-FG corpus (\secref{sec:dataset}) as the target domain examples. Note that $M$ is a minimal number (namely, $M=5$) in our setting, and the task has an aspect of low-resource domain adaptation~\citep{xu2021gradual}. Note also that training with only the \revise{cooking} or target-domain examples becomes zero-shot or few-shot learning scenarios, respectively. 

\begin{table*}[t]
    \centering
    \scalebox{0.985}{
    \begin{tabular}{l|l}\hline
        Domain & Task examples \\\hline
        \textit{Food and Entertaining} & \textit{Cooking acorn squash}, \textit{Making lavender tea}, \textit{Baking a cherry pie} \\
        \textit{Hobbies and Crafts} & \textit{Making a bar soap}, \textit{Making a duct tape bow}, \textit{Making a paper box} \\
        \textit{Home and Garden} & \textit{Cleaning a mattress pad}, \textit{Installing a microwave}, \textit{Making a scented candle} \\
        \textit{Cars \& Other Vehicles} & \textit{Fixing a slipped bike chain}, \textit{Cleaning car window}, \textit{Cleaning tail lights} \\\hline
    \end{tabular}
    }
    \caption{Examples of article titles for each domain.}
    \label{tab:article_title_examples}
\end{table*}
\begin{table*}[t]
    \centering
    \begin{tabular}{l|ccccc} \hline
        Domain & \# Characters & \# Words & \# Steps & \# Tags & \# Labels \\\hline
        \textit{Food and Entertaining} & 10,167 & 2,761 & 224 & 1,132 & 1,124 \\
        \textit{Hobbies and Crafts} & ~~9,407 & 2,556 & 247 & 1,062 & 1,076 \\
        \textit{Home and Garden} & ~~7,700 & 2,010 & 205 & ~~~894 & ~~~886 \\
        \textit{Cars \& Other Vehicles} & ~~6,432 & 1,622 & 173 & ~~~625 & ~~~622 \\\hline
    \end{tabular}
    \caption{Statistics of the wikiHow-FG corpus.}
    \label{tab:stats}
\end{table*}
\begin{figure}[t]
    \centering
    \includegraphics[scale=0.35]{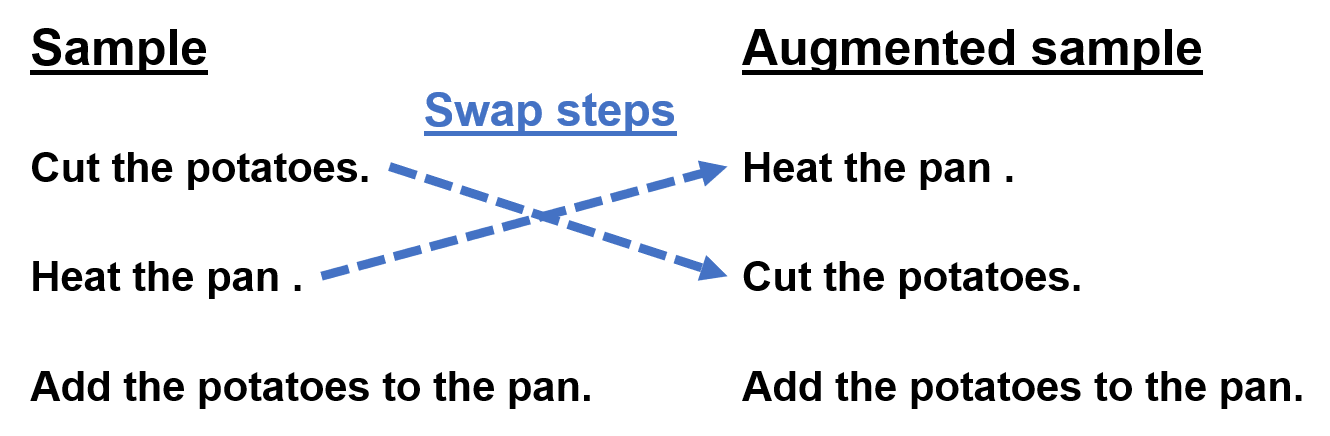}
    \caption{Example of swapping steps. The first and second steps are swappable without violating causality.}
    \label{fig:step_swaps_example}
\end{figure}
\subsection{Data augmentation}\label{subsec:data_augmentation}
For improving performance in a low-resource setting, data augmentation is one of possible solutions~\citep{fadaee2017data,ding2020daga}. In this work, we consider the following two data augmentation techniques: step swapping and word replacement.

\paragraph{Step swapping} augments an example by replacing two arbitrary steps in an article as illustrated in \figref{fig:step_swaps_example}. However, randomly choosing and swapping two steps might break their causal relationship. For example, we cannot swap two steps ``Cut the potatoes.'' and ``Add the potatoes to the pan.'' in \figref{fig:step_swaps_example}. \revise{In this work, we augment examples keeping this constraint by using flow graph annotations.}

\paragraph{Word replacement} augments an example by replacing a word with an arbitrary one. For each word in a step, we replace a word with one of its synonyms from WordNet~\citep{daiadel2020analysis} with a probability $p$. 

\section{wikiHow-FG corpus}\label{sec:dataset}
The wikiHow-FG corpus is a new flow graph corpus from articles on wikiHow\footnote{\url{https://www.wikihow.com}}, a website with more than 110K how-to articles. wikiHow articles have been used as a language resource for procedural texts~\citep{zhou2019learning,zellers2019hellaswag,zhang2020reasoning,zhang2020intent,zhou2022show}. In the following, we describe the data collection, annotation procedure, and statistics of the annotation results.

\subsection{Data collection}\label{subsec:data_collection}
For the target domains, we selected four categories from wikiHow: (i) \textit{Food and Entertaining}, (ii) \textit{Hobbies and Crafts}, (iii) \textit{Home and Garden}, and (iv) \textit{Cars \& Other Vehicles}. We decided on those domains because many articles target and interact with substantial objects. We show examples of article titles in \tabref{tab:article_title_examples}. \textit{Food and Entertaining} is a domain close to the cooking domain as with the English-FG. \textit{Hobbies and Crafts} is far from the cooking domain in the sense that they assemble non-edible materials (e.g., \textit{making a bar soap}). The remaining two domains are further from these domains because they contain non-assembly tasks (e.g., \textit{cleaning a table} and \textit{fixing a broken chain}).

We collected $30$ articles for each domain from the wikiHow corpus~\citep{zhang2020reasoning}. To exclude low-quality articles, we collected articles with 25 or more words and more than 50\% user ratings. We then manually excluded articles with too abstract goals or not targeting substantial objects. We used article headlines as steps for annotation and experiments. Note that each article has a title and visual information (e.g., images or videos) describing the procedures. Exploiting them is interesting, but we leave that direction for future work.

\subsection{Annotation procedure}\label{subsec:annotation_procedure}
Due to the dense, complex nature of the flow graph, constructing a high-quality corpus is a challenging problem. In order to guarantee the annotation quality, we first trained an annotator with $10$ recipes sampled from the English-FG corpus~\citep{yamakata2020english}. The training continued until the inter-annotator agreements with the ground-truth annotations reached over 80\%. We then asked the annotator to annotate wikiHow articles. For both the node and edge annotations, we used the flow graph annotation tool~\citep{shirai2022visual}.\footnote{Before the annotation, we tokenized steps into words with the stanza toolkit~\citep{qi2020stanza}.} The whole annotation took $40$ hours.

\subsection{Statistics}\label{subsec:stats}
We show statistics of the wikiHow corpus in \tabref{tab:stats}. The articles comprise $280.9$ characters, $74.6$ words, and $7.08$ steps on average. The average number of tags and labels per article is $37.73$ ($\pm 16.48$) and $37.47$ ($\pm 17.21$) on \textit{Food and Entertaining}, $35.40$ ($\pm 9.08$) and $35.87$ ($\pm 9.44$) on \textit{Hobbies and Crafts}, $29.80$ ($\pm 7.83$) and $29.53$ ($\pm 8.29$) on \textit{Home and Garden}, and $20.83$ ($\pm 6.40$) and $20.73$ ($\pm 7.14$) on \textit{Cars \& Other Vehicles}.
Since articles on \textit{Home and Garden} and \textit{Cars \& Other Vehicles} have fewer words than the other two domains, the number of tags and labels also becomes smaller. 

\begin{table}[t]
    \centering
    \begin{tabular}{l|c} \hline
        Annotation type & Agreement \\\hline
        Node annotation & 90.17\% \\
        Edge annotation & 57.43\% \\\hline
    \end{tabular}
    \caption{Inter-annotator agreements.}
    \label{tab:annotator_agreements}
\end{table}
\subsection{Inter-annotator agreements}
To assess the consistency of the annotations, we asked another annotator to re-annotate 10\% of articles for each domain. We then measured F1 scores between the two sets of annotations using the original annotations as the ground-truth ones. \tabref{tab:annotator_agreements} shows the results. The agreement for the node annotation (90.14\%) was high, considering entities corresponding to the tags greatly change depending on the domain. For the edge annotation, the agreement (57.43\%) drops from the one for node annotation. However, this agreement is also high considering errors from the node annotation influence this step and that a large number of candidate edges for the annotation.

\begin{table*}[t]
    \centering
    \begin{tabular}{l|c|cc|ccc} \hline
        \multirow{2}{*}{Domain} & \multirow{2}{*}{Model} & \multicolumn{2}{c|}{Augmentation} & \multirow{2}{*}{Prec.} & \multirow{2}{*}{Recall} & \multirow{2}{*}{F1} \\\cline{3-4}
        & & Step-swap & Word-replace & \\\hline
        
        \multirow{5}{*}{\textit{Food and Entertaining}} 
        & Target-only & & & 0.770 & 0.784 & 0.777 \\
        & Cooking-only & & & 0.884 & 0.877 & 0.880 \\
        & Domain-adaptation & & & 0.890 & 0.892 & 0.891 \\
        & Domain-adaptation & \checkmark & & \textbf{0.894} & \textbf{0.895} & \textbf{0.895} \\
        & Domain-adaptation & & \checkmark & 0.885 & 0.891 & 0.888 \\\hline
        
        \multirow{5}{*}{\textit{Hobbies and Crafts}} 
        & Target-only & & & 0.698 & 0.707 & 0.702 \\
        & Cooking-only & & & 0.703 & 0.684 & 0.693 \\
        & Domain-adaptation & & & \textbf{0.794} & \textbf{0.805} & \textbf{0.799} \\
        & Domain-adaptation & \checkmark & & 0.784 & 0.795 & 0.789 \\
        & Domain-adaptation & & \checkmark & 0.781 & 0.790 & 0.785 \\\hline
        
        \multirow{5}{*}{\textit{Home and Garden}} 
        & Target-only & & & 0.663 & 0.676 & 0.669 \\
        & Cooking-only & & & 0.734 & 0.742 & 0.738 \\
        & Domain-adaptation & & & 0.780 & 0.786 & 0.783 \\
        & Domain-adaptation & \checkmark & & \textbf{0.787} & \textbf{0.791} & \textbf{0.786} \\
        & Domain-adaptation & & \checkmark & 0.765 & 0.773 & 0.769 \\\hline
        
        \multirow{5}{*}{\textit{Cars \& Other Vehicles}} 
        & Target-only & & & 0.650 & 0.690 & 0.669 \\
        & Cooking-only & & & 0.646 & 0.695 & 0.670 \\
        & Domain-adaptation & & & \textbf{0.748} & \textbf{0.784} & \textbf{0.765} \\
        & Domain-adaptation & \checkmark & & 0.734 & \textbf{0.784} & 0.761 \\
        & Domain-adaptation & & \checkmark & 0.729 & 0.772 & 0.750 \\\hline
    \end{tabular}
    \caption{Results of the node prediction experiments. The check mark symbol ($\checkmark$) indicates the used training data (in cooking and target domains) and augmentation techniques (step-swap and word-replace).}
    \label{tab:ner_results}
\end{table*}
\section{Node prediction}\label{sec:node_prediction}
\subsection{Experimental settings}\label{subsec:node_exp_setting}
\paragraph{Model.} 
We adopted a BiLSTM-CRF model~\citep{lample2016neural} for node prediction by replacing a BiLSTM encoder with a pre-trained language model (LM)~\citep{devlin2019bert,liu2019roberta,he2021deberta}. While previous work~\citep{yamakata2020english} used a pre-trained BERT~\citep{devlin2019bert} as the encoder, we used a pre-trained DeBERTa~\citep{he2021deberta}.\footnote{In our preliminary experiments, we confirmed that DeBERTa improves the accuracy of BERT by 0.47\% on the English-FG corpus.} This model has 140M parameters in total.

\paragraph{Training.} 
We trained a domain adaptation model by first training on the cooking domain data (English-FG) and then training again on the target domain data (wikiHow-FG). Note that we use only target domain examples of the wikiHow-FG corpus (e.g., when targeting \textit{Hobbies and Crafts}, we do not use examples in the other three domains.). We also train models only on the cooking or target domain data and report the results to compare with the domain adaptation results.

For an optimization method, we used AdamW~\citep{loshchilov2019decoupled} with an initial learning rate of $5.0 \times 10^{-5}$ and a weight decay of $1.0 \times 10^{-5}$. We tuned a learning rate with a cosine-annealing~\citep{loshchilov2019decoupled} ($S_{\text{d}}$ steps) with a linear warm-up ($S_{\text{w}}$ steps) at every iteration. We created a mini-batch from $B$ articles. We set $(B, S_{\text{w}}, S_{\text{d}}) = (5, 500, 4500)$ and $(B, S_{\text{w}}, S_{\text{d}}) = (3, 100, 900)$ for training on the English-FG corpus and the wikiHow-FG corpus, respectively. We tuned these hyper-parameters on the development set. We used the data augmentation techniques only for the target domain data. For the step swapping, we created $5$ augmented examples at maximum from one example. For the word replacement, we set $0.5$ to $p$ and created $10$ examples from one example. 

\paragraph{Evaluation.} 
We split the English-FG corpus into $80\%$ for training, $10\%$ for validation, and the rest of $10\%$ for testing. For the wikiHow-FG corpus, we split $30$ articles of each domain into $6$ folds. For more reliable results, we performed $6$-fold cross-validation by using $1$ fold for training, $1$ fold for validation, and the remaining $4$ folds for testing. We used precision, recall, and F1 for evaluation metrics, following \citet{yamakata2020english}. On the evaluation, we report the average scores of the models on the test set. 

\paragraph{Model configurations.}
We refer to the domain adaptation models as \textbf{domain-adaptation} models. We also refer to the models trained only on the English-FG or the wikiHow-FG corpus as \textbf{cooking-only} and \textbf{target-only} models, respectively. 

\begin{table*}[t]
    \centering
    \scalebox{0.86}{
    \begin{tabular}{l|ccc|ccc|ccc} \hline
        \multirow{3}{*}{Domain} & \multicolumn{3}{c|}{\textsf{Ae}} & \multicolumn{3}{c|}{\textsf{C}} & \multicolumn{3}{c}{\textsf{T}} \\\cline{2-10}
        
        & \multirow{2}{*}{Duplicates} & \multicolumn{2}{c|}{F1} & \multirow{2}{*}{Duplicates} & \multicolumn{2}{c|}{F1} & \multirow{2}{*}{Duplicates} & \multicolumn{2}{c}{F1} \\\cdashline{3-4}[2.5pt/1.0pt]\cdashline{6-7}[2.5pt/1.0pt]\cdashline{9-10}[2.5pt/1.0pt]
        & & Src & Adpt & & Src & Adpt & & Src & Adpt \\\hline
        
        \textit{Food and Entertaining} & 92.06\% & 0.941 & 0.952 & 72.11\% & 0.932 & 0.933 & 77.94\% & 0.896 & 0.882 \\
        \textit{Hobbies and Crafts} & 69.03\% & 0.943 & 0.951 & 10.33\% & 0.717 & 0.833 & 51.79\% & 0.398 & 0.588 \\
        \textit{Home and Garden} & 65.19\% & 0.954 & 0.961 & 18.40\% & 0.716 & 0.795 & 43.55\% & 0.567 & 0.678 \\
        \textit{Cars \& Other Vehicles} & 46.04\% & 0.905 & 0.919 & ~~6.88\% & 0.666 & 0.805 & 27.47\% & 0.459 & 0.557 \\\hline
    \end{tabular}
    }
    \caption{F1 scores of \textsf{Ae}, \textsf{C}, and \textsf{T} tags with the percentage of entities that appeared in the English-FG corpus and also in the wikiHow-FG corpus. Src and Adpt denote \textbf{cooking-only} and \textbf{domain-adaptation} models, respectively.}
    \label{tab:ner_tag_results}
\end{table*}

\begin{table*}[t]
    \centering
    \begin{tabular}{l|c|cc|ccc} \hline
        \multirow{2}{*}{Domain} & \multirow{2}{*}{Model} & \multicolumn{2}{c|}{Augmentation} & \multirow{2}{*}{Prec.} & \multirow{2}{*}{Recall} & \multirow{2}{*}{F1} \\\cline{3-4}
        & & Step-Swap & Word-Replace & \\\hline
        
        \multirow{5}{*}{\textit{Food and Entertaining}} 
        & Target-only & & & 0.335 & 0.338 & 0.337 \\
        & Cooking-only & & & 0.725 & 0.731 & 0.728 \\
        & Domain-adaptation & & & 0.750 & \textbf{0.756} & \textbf{0.753} \\
        & Domain-adaptation & \checkmark & & 0.747 & 0.752 & 0.750 \\
        & Domain-adaptation & & \checkmark & \textbf{0.761} & 0.752 & 0.749 \\\hline
        
        \multirow{5}{*}{\textit{Hobbies and Crafts}} 
        & Target-only & & & 0.285 & 0.281 & 0.283 \\
        & Cooking-only & & & 0.613 & 0.605 & 0.609 \\
        & Domain-adaptation & & & 0.649 & 0.640 & 0.644 \\
        & Domain-adaptation & \checkmark & & 0.646 & 0.638 & 0.642 \\
        & Domain-adaptation & & \checkmark & \textbf{0.653} & \textbf{0.644} & \textbf{0.648} \\\hline
        
        \multirow{5}{*}{\textit{Home and Garden}} 
        & Target-only & & & 0.229 & 0.232 & 0.231 \\
        & Cooking-only & & & 0.644 & 0.649 & 0.646 \\
        & Domain-adaptation & & & 0.659 & 0.665 & 0.662 \\
        & Domain-adaptation & \checkmark & & 0.656 & 0.662 & 0.659 \\
        & Domain-adaptation & & \checkmark & \textbf{0.674} & \textbf{0.680} & \textbf{0.677} \\\hline
        
        \multirow{5}{*}{\textit{Cars \& Other Vehicles}} 
        & Target-only & & & 0.154 & 0.155 & 0.154 \\
        & Cooking-only & & & 0.587 & 0.590 & 0.587 \\
        & Domain-adaptation & & & 0.607 & 0.610 & 0.609 \\
        & Domain-adaptation & \checkmark & & 0.607 & 0.610 & 0.608 \\
        & Domain-adaptation & & \checkmark & \textbf{0.617} & \textbf{0.620} & \textbf{0.618} \\\hline
    \end{tabular}
    \caption{Results of the edge prediction experiments. The check mark symbol ($\checkmark$) indicates the used training data (in cooking and target domains) and augmentation techniques (step-swap and word-replace).}
    \label{tab:parsing_results}
\end{table*}
\subsection{Results}\label{subsec:ner_results}
The results are shown in \tabref{tab:ner_results}. We see that the \textbf{target-only} models achieve an F1 score of 66.9\% or more in all the target domains. This implies that the node prediction model can predict nodes to an extent with a few annotated articles. We also see that the \textbf{cooking-only} models achieve competitive performance with the \textbf{target-only} ones and outperform them in the two domains of \textit{Food and Entertaining} and \textit{Home and Garden}. Particularly in \textit{Food and Entertaining}, the \textbf{cooking-only} model surpasses the \textbf{target-only} one by 10.3\% in F1. We consider that this domain is close to the cooking domain of the English-FG; thus, the \textbf{cooking-only} model is more advantageous as it can access more examples.

Next, the \textbf{domain-adaptation} models achieve the best performance in all the domains compared with the \textbf{cooking-only} and \textbf{target-only} models (76.5\% or more F1). The most significant improvements are obtained in the two domains of \textit{Hobbies and Crafts} and \textit{Cars \& Other Vehicles} (9.5\% and 10.5\% improvements in F1). These results indicate that domain adaptation from the cooking to the target domain is effective for training the node prediction model.

Third, we see that using the augmented data by the step swapping slightly improves performance from the \textbf{domain-adaptation} models in \textit{Food and Entertaining} and \textit{Home and Garden}. On the other hand, the word replacement does not contribute to any improvement. One possible reason is that a replaced word does not necessarily match the corresponding tag, and this disrupts the improvement. 

\subsection{Tag-level prediction performance}
Entities for each tag can greatly change depending on the domain. In that case, the degree of improvement from the \textbf{cooking-only} to the \textbf{domain-adaptation} model is expected to increase as the duplicate entities between the domains decrease. To investigate this assumption, we measured tag-level prediction performance in F1 with the duplicate ratio of entities of the wikiHow-FG in the English-FG. We targeted the three tags of \textsf{Ae}, \textsf{C}, and \textsf{T} because these tags frequently appear in all the domains. 

The results are shown in \tabref{tab:ner_tag_results}. For \textsf{Ae}, the degree of improvement from the \textbf{cooking-only} to the \textbf{domain-adaptation} model is small regardless of the duplicate ratios, which is contrary to our assumption. These results imply that recognizing entities for \textsf{Ae} is easy irrespective of the domain. For \textsf{C} and \textsf{T}, the \textbf{domain-adaptation} models significantly outperform the \textbf{cooking-only} ones in the three domains other than \textit{Food and Entertaining}. These results imply that the domain adaptation is effective for recognizing \textsf{C} and \textsf{T} tags when the domain is further from the cooking domain.

\section{Edge prediction}\label{sec:edge_prediction}
\subsection{Experimental settings}
\paragraph{Model.}
We adopted a biaffine dependency parser~\citep{dozat2018simpler} for edge prediction.\footnote{Previous work~\citep{maeta2015framework} used a linear model, but we confirmed that our model achieves higher performance on the English-FG corpus.} This model uses separate modules for edge prediction and label prediction. The resulting loss $l$ is defined as a weighted sum of losses from the two modules:
\begin{equation}
    l = \lambda l^{\text{(edge)}} + (1 - \lambda) l^{\text{(label)}},
\end{equation}
where $\lambda$ controls the strength of the two losses. We empirically set 0.5 to $\lambda$. We used a pre-trained DeBERTa~\citep{he2021deberta} to obtain contextualized word representations. This model has 149M parameters in total. 

\paragraph{Training.}
Similarly to \secref{subsec:node_exp_setting}, we trained a domain adaptation model for edge prediction first on the English-FG corpus and then on the wikiHow-FG corpus. For an optimization method, we used AdamW~\citep{loshchilov2019decoupled} with a combination of a cosine-annealing and linear warm-up learning rate scheduling method. We used the same hyperparameters in \secref{subsec:node_exp_setting}.

\paragraph{Evaluation.}
We used the same splits of the English-FG and wikiHow-FG corpora as in \secref{sec:node_prediction} and performed $6$-fold cross-validation for more reliable results. We report the average scores of the models on the test set. For evaluation metrics, we used precision, recall, and F1 between predicted and ground-truth labeled edges of $(u, v, l)$. 

\paragraph{Model configurations.}
We used the same model notations of \textbf{cooking-only}, \textbf{target-only}, and \textbf{domain-adaptation} models as in \secref{sec:node_prediction}.

\subsection{Results}\label{subsec:edge_results}
The results are shown in \tabref{tab:parsing_results}. We used ground-truth tags to identify nodes. Contrary to the node prediction results, the \textbf{target-only} models achieve poor performance in all the domains (33.8\% or less in all the metrics). On the other hand, the scores of the \textbf{cooking-only} models are more than twice those of the \textbf{target-only} models. These results show that the edge prediction model requires more training examples than the node prediction one. These also show that with the English-FG corpus, predicting edges with 58.7\% or more F1 is possible in non-cooking domains.

Next, the \textbf{domain-adaptation} models outperform the \textbf{target-only} and \textbf{cooking-only} ones in all the domains. This is consistent with the results in the node prediction task. These results mean that the domain adaptation from the cooking to the target domain is also effective for the edge prediction model. For the results with the data augmentation techniques, the step swapping does not contribute to any improvement, contrary to \secref{subsec:ner_results}. The word replacement improves the performance of the \textbf{domain-adaptation} models in the three domains other than \textit{Food and Entertaining}.

\begin{table}[t]
    \centering
    \begin{tabular}{l|c} \hline
        Domain & F1 \\\hline
        \textit{Food and Entertaining} & 0.679 ~~(-9.8\%) \\ 
        \textit{Hobbies and Crafts} & 0.501 (-22.2\%) \\
        \textit{Home and Garden} & 0.494 (-25.4\%) \\
        \textit{Cars \& Other Vehicles} & 0.449 (-26.3\%) \\\hline
    \end{tabular}
    \caption{Results of the pipeline experiments. The inside of the parenthesis represents the performance drop from the \textbf{domain-adaptation} model with ground-truth tags.}
    \label{tab:parsing_pipeline_results}
\end{table}
\subsection{Pipeline experiments}
So far, the model has used ground-truth tags to identify nodes. However, in a realistic scenario, the model must predict labeled edges with the predicted nodes. In this scenario, errors in the node prediction step would affect performance in the edge prediction step. To investigate edge prediction performance in this setting, we conducted experiments of edge prediction with the predicted nodes. We predicted nodes using the models in \secref{subsec:ner_results}. In order to evaluate the model with tag information, we measured F1 of tuples of $(u, v, l, n_u, n_v)$ between ground-truth and predicted ones, where $n_u$ and $n_v$ are the tags of the starting and ending nodes, respectively.

\tabref{tab:parsing_pipeline_results} shows the results with performance drops from those in \tabref{tab:parsing_results}. We see that 9.8\% drops in \textit{Food and Entertaining}, and more significant drops occur in the other three domains (about 24.6\%). In these three domains, F1 scores of the node prediction are about 10\% smaller than that of \textit{Food and Entertaining}, and this gap would cause such large performance drops. We consider that improving node prediction performance would alleviate these drops.

\section{Related work}
\citet{mori2014flow} designed a flow graph (FG) representation in the cooking domain and introduced a corpus of recipe texts. Subsequent works introduced a corpus in English~\citep{yamakata2020english} and corpora with visual annotations~\citep{nishimura2020visual,shirai2022visual}. \citet{maeta2015framework} proposed a method for an automatic FG prediction. Our work stems from this line of research and is the first attempt to apply the framework to non-cooking domains. Ours is also the first work to use a neural network-based method for the edge prediction.

Other than the recipe FG, there are several works that focus on obtaining an actionable representation from a procedural text. In cooking, \citet{kiddon2015mise} proposed an unsupervised EM algorithm, while \citet{pan2020multimodal,papadopoulos2022learning} proposed supervised approaches. In biochemistry, \citet{kulkarni2018annotated,tamari2021process} introduced datasets for mapping wet lab protocols to an action graph. In material science, \citet{kuniyoshi2020annotating} represented the synthesis process with flow graphs. The works \citep{pan2020multimodal,papadopoulos2022learning,tamari2021process,kuniyoshi2020annotating} are especially close to ours in the sense that they aim to obtain a document-level action graph in a supervised way.

\section{Conclusion}
We proposed a framework based on the English-FG and investigated flow graph prediction performance in non-cooking domains. We presented the wikiHow-FG corpus from wikiHow articles. We considered domain adaptation from the cooking to the target domain. Experimental results show that domain adaptation models outperform those trained only on the cooking or target domain data. In future work, we consider applying this framework to other domains, such as material science and biochemistry. One can also try improving performance using more sophisticated data augmentation techniques. We hope that our work will provide new insights into procedural text understanding.

\section*{Acknowledgments}
We would like to thank anonymous reviewers for their insightful comments. This work was supported by JSPS KAKENHI Grant Number 20H04210 and 21H04910.

\bibliographystyle{acl_natbib}
\bibliography{anthology,custom}

\begin{thebibliography}{31}
\expandafter\ifx\csname natexlab\endcsname\relax\def\natexlab#1{#1}\fi

\bibitem[{Bollini et~al.(2013)Bollini, Tellex, Thompson, Roy, and
  Rus}]{bollini2013interpreting}
Mario Bollini, Stefanie Tellex, Tyler Thompson, Nicholas Roy, and Daniela Rus.
  2013.
\newblock \href {https://doi.org/10.1007/978-3-319-00065-7_33} {Interpreting
  and executing recipes with a cooking robot}.
\newblock In \emph{Experimental Robotics}, pages 481--495. Springer.

\bibitem[{Dai and Adel(2020)}]{daiadel2020analysis}
Xiang Dai and Heike Adel. 2020.
\newblock \href {https://doi.org/10.18653/v1/2020.coling-main.343} {An analysis
  of simple data augmentation for named entity recognition}.
\newblock In \emph{Proceedings of the 28th International Conference on
  Computational Linguistics}, pages 3861--3867. International Committee on
  Computational Linguistics.

\bibitem[{Devlin et~al.(2019)Devlin, Chang, Lee, and
  Toutanova}]{devlin2019bert}
Jacob Devlin, Ming-Wei Chang, Kenton Lee, and Kristina Toutanova. 2019.
\newblock \href {https://doi.org/10.18653/v1/N19-1423} {{BERT}: Pre-training of
  deep bidirectional transformers for language understanding}.
\newblock In \emph{Proceedings of the 2019 Conference of the North {A}merican
  Chapter of the Association for Computational Linguistics: Human Language
  Technologies (Volume 1: Long and Short Papers)}, pages 4171--4186.
  Association for Computational Linguistics.

\bibitem[{Ding et~al.(2020)Ding, Liu, Bing, Kruengkrai, Nguyen, Joty, Si, and
  Miao}]{ding2020daga}
Bosheng Ding, Linlin Liu, Lidong Bing, Canasai Kruengkrai, Thien~Hai Nguyen,
  Shafiq Joty, Luo Si, and Chunyan Miao. 2020.
\newblock \href {https://doi.org/10.18653/v1/2020.emnlp-main.488} {{DAGA}: Data
  augmentation with a generation approach for low-resource tagging tasks}.
\newblock In \emph{Proceedings of the 2020 Conference on Empirical Methods in
  Natural Language Processing (EMNLP)}, pages 6045--6057. Association for
  Computational Linguistics.

\bibitem[{Dozat and Manning(2018)}]{dozat2018simpler}
Timothy Dozat and Christopher~D. Manning. 2018.
\newblock \href {https://doi.org/10.18653/v1/P18-2077} {Simpler but more
  accurate semantic dependency parsing}.
\newblock In \emph{Proceedings of the 56th Annual Meeting of the Association
  for Computational Linguistics (Volume 2: Short Papers)}, pages 484--490.
  Association for Computational Linguistics.

\bibitem[{Fadaee et~al.(2017)Fadaee, Bisazza, and Monz}]{fadaee2017data}
Marzieh Fadaee, Arianna Bisazza, and Christof Monz. 2017.
\newblock \href {https://doi.org/10.18653/v1/P17-2090} {Data augmentation for
  low-resource neural machine translation}.
\newblock In \emph{Proceedings of the 55th Annual Meeting of the Association
  for Computational Linguistics (Volume 2: Short Papers)}, pages 567--573.
  Association for Computational Linguistics.

\bibitem[{He et~al.(2021)He, Liu, Gao, and Chen}]{he2021deberta}
Pengcheng He, Xiaodong Liu, Jianfeng Gao, and Weizhu Chen. 2021.
\newblock \href {https://openreview.net/forum?id=XPZIaotutsD} {Deberta:
  decoding-enhanced bert with disentangled attention}.
\newblock In \emph{9th International Conference on Learning Representations}.

\bibitem[{Kiddon et~al.(2015)Kiddon, Ponnuraj, Zettlemoyer, and
  Choi}]{kiddon2015mise}
Chlo{\'e} Kiddon, Ganesa~Thandavam Ponnuraj, Luke Zettlemoyer, and Yejin Choi.
  2015.
\newblock \href {https://doi.org/10.18653/v1/D15-1114} {Mise en place:
  Unsupervised interpretation of instructional recipes}.
\newblock In \emph{Proceedings of the 2015 Conference on Empirical Methods in
  Natural Language Processing}, pages 982--992. Association for Computational
  Linguistics.

\bibitem[{Kulkarni et~al.(2018)Kulkarni, Xu, Ritter, and
  Machiraju}]{kulkarni2018annotated}
Chaitanya Kulkarni, Wei Xu, Alan Ritter, and Raghu Machiraju. 2018.
\newblock \href {https://doi.org/10.18653/v1/N18-2016} {An annotated corpus for
  machine reading of instructions in wet lab protocols}.
\newblock In \emph{Proceedings of the 2018 Conference of the North {A}merican
  Chapter of the Association for Computational Linguistics: Human Language
  Technologies, Volume 2 (Short Papers)}, pages 97--106. Association for
  Computational Linguistics.

\bibitem[{Kuniyoshi et~al.(2020)Kuniyoshi, Makino, Ozawa, and
  Miwa}]{kuniyoshi2020annotating}
Fusataka Kuniyoshi, Kohei Makino, Jun Ozawa, and Makoto Miwa. 2020.
\newblock \href {https://aclanthology.org/2020.lrec-1.239} {Annotating and
  extracting synthesis process of all-solid-state batteries from scientific
  literature}.
\newblock In \emph{Proceedings of the Twelfth Language Resources and Evaluation
  Conference}, pages 1941--1950. European Language Resources Association.

\bibitem[{Lample et~al.(2016)Lample, Ballesteros, Subramanian, Kawakami, and
  Dyer}]{lample2016neural}
Guillaume Lample, Miguel Ballesteros, Sandeep Subramanian, Kazuya Kawakami, and
  Chris Dyer. 2016.
\newblock \href {https://doi.org/10.18653/v1/N16-1030} {Neural architectures
  for named entity recognition}.
\newblock In \emph{Proceedings of the 2016 Conference of the North {A}merican
  Chapter of the Association for Computational Linguistics: Human Language
  Technologies}, pages 260--270. Association for Computational Linguistics.

\bibitem[{Liu et~al.(2019)Liu, Ott, Goyal, Du, Joshi, Chen, Levy, Lewis,
  Zettlemoyer, and Stoyanov}]{liu2019roberta}
Yinhan Liu, Myle Ott, Naman Goyal, Jingfei Du, Mandar Joshi, Danqi Chen, Omer
  Levy, Mike Lewis, Luke Zettlemoyer, and Veselin Stoyanov. 2019.
\newblock \href {https://doi.org/10.48550/ARXIV.1907.11692} {Roberta: A
  robustly optimized bert pretraining approach}.
\newblock \emph{arXiv preprint arXiv:1907.11692}.

\bibitem[{Loshchilov and Hutter(2019)}]{loshchilov2019decoupled}
Ilya Loshchilov and Frank Hutter. 2019.
\newblock \href {https://openreview.net/forum?id=Bkg6RiCqY7} {Decoupled weight
  decay regularization}.
\newblock In \emph{Proceedings of the 7th International Conference on Learning
  Representations}.

\bibitem[{Maeta et~al.(2015)Maeta, Sasada, and Mori}]{maeta2015framework}
Hirokuni Maeta, Tetsuro Sasada, and Shinsuke Mori. 2015.
\newblock \href {https://doi.org/10.18653/v1/W15-2206} {A framework for
  procedural text understanding}.
\newblock In \emph{Proceedings of the 14th International Conference on Parsing
  Technologies}, pages 50--60. Association for Computational Linguistics.

\bibitem[{McDonald et~al.(2005)McDonald, Pereira, Ribarov, and
  Haji{\v{c}}}]{mcdonald2005non}
Ryan McDonald, Fernando Pereira, Kiril Ribarov, and Jan Haji{\v{c}}. 2005.
\newblock \href {https://aclanthology.org/H05-1066} {Non-projective dependency
  parsing using spanning tree algorithms}.
\newblock In \emph{Proceedings of Human Language Technology Conference and
  Conference on Empirical Methods in Natural Language Processing}, pages
  523--530. Association for Computational Linguistics.

\bibitem[{Momouchi(1980)}]{momouchi1980control}
Yoshio Momouchi. 1980.
\newblock \href {https://aclanthology.org/C80-1016} {Control structures for
  actions in procedural texts and {PT}-chart}.
\newblock In \emph{{COLING} 1980 Volume 1: The 8th International Conference on
  Computational Linguistics}.

\bibitem[{Mori et~al.(2014)Mori, Maeta, Yamakata, and Sasada}]{mori2014flow}
Shinsuke Mori, Hirokuni Maeta, Yoko Yamakata, and Tetsuro Sasada. 2014.
\newblock \href {https://aclanthology.org/L14-1594} {Flow graph corpus from
  recipe texts}.
\newblock In \emph{Proceedings of the Ninth International Conference on
  Language Resources and Evaluation}, pages 2370--2377.

\bibitem[{Nishimura et~al.(2020)Nishimura, Tomori, Hashimoto, Hashimoto,
  Yamakata, Harashima, Ushiku, and Mori}]{nishimura2020visual}
Taichi Nishimura, Suzushi Tomori, Hayato Hashimoto, Atsushi Hashimoto, Yoko
  Yamakata, Jun Harashima, Yoshitaka Ushiku, and Shinsuke Mori. 2020.
\newblock \href {https://aclanthology.org/2020.lrec-1.527} {Visual grounding
  annotation of recipe flow graph}.
\newblock In \emph{Proceedings of the 12th Language Resources and Evaluation
  Conference}, pages 4275--4284. European Language Resources Association.

\bibitem[{Pan et~al.(2020)Pan, Chen, Wu, Liu, Ngo, Kan, Jiang, and
  Chua}]{pan2020multimodal}
Liang-Ming Pan, Jingjing Chen, Jianlong Wu, Shaoteng Liu, Chong-Wah Ngo,
  Min-Yen Kan, Yugang Jiang, and Tat-Seng Chua. 2020.
\newblock \href {https://doi.org/10.1145/3394171.3413765} {Multi-modal cooking
  workflow construction for food recipes}.
\newblock In \emph{Proceedings of the 28th ACM International Conference on
  Multimedia}, page 1132–1141. Association for Computing Machinery.

\bibitem[{Papadopoulos et~al.(2022)Papadopoulos, Mora, Chepurko, Huang, Ofli,
  and Torralba}]{papadopoulos2022learning}
Dim~P. Papadopoulos, Enrique Mora, Nadiia Chepurko, Kuan~Wei Huang, Ferda Ofli,
  and Antonio Torralba. 2022.
\newblock \href
  {https://openaccess.thecvf.com/content/CVPR2022/html/Papadopoulos_Learning_Program_Representations_for_Food_Images_and_Cooking_Recipes_CVPR_2022_paper.html}
  {Learning program representations for food images and cooking recipes}.
\newblock In \emph{Proceedings of the IEEE/CVF Conference on Computer Vision
  and Pattern Recognition (CVPR)}, pages 16559--16569.

\bibitem[{Qi et~al.(2020)Qi, Zhang, Zhang, Bolton, and Manning}]{qi2020stanza}
Peng Qi, Yuhao Zhang, Yuhui Zhang, Jason Bolton, and Christopher~D. Manning.
  2020.
\newblock \href {https://doi.org/10.18653/v1/2020.acl-demos.14} {{S}tanza: A
  python natural language processing toolkit for many human languages}.
\newblock In \emph{Proceedings of the 58th Annual Meeting of the Association
  for Computational Linguistics: System Demonstrations}, pages 101--108.
  Association for Computational Linguistics.

\bibitem[{Ramshaw and Marcus(1995)}]{ramshaw1995text}
Lance Ramshaw and Mitch Marcus. 1995.
\newblock \href {https://aclanthology.org/W95-0107} {Text chunking using
  transformation-based learning}.
\newblock In \emph{Third Workshop on Very Large Corpora}.

\bibitem[{Shirai et~al.(2022)Shirai, Hashimoto, Nishimura, Kameko, Kurita,
  Ushiku, and Mori}]{shirai2022visual}
Keisuke Shirai, Atsushi Hashimoto, Taichi Nishimura, Hirotaka Kameko, Shuhei
  Kurita, Yoshitaka Ushiku, and Shinsuke Mori. 2022.
\newblock \href {https://aclanthology.org/2022.coling-1.315} {Visual recipe
  flow: A dataset for learning visual state changes of objects with recipe
  flows}.
\newblock In \emph{Proceedings of the 29th International Conference on
  Computational Linguistics}, pages 3570--3577. International Committee on
  Computational Linguistics.

\bibitem[{Tamari et~al.(2021)Tamari, Bai, Ritter, and
  Stanovsky}]{tamari2021process}
Ronen Tamari, Fan Bai, Alan Ritter, and Gabriel Stanovsky. 2021.
\newblock \href {https://doi.org/10.18653/v1/2021.eacl-main.187} {Process-level
  representation of scientific protocols with interactive annotation}.
\newblock In \emph{Proceedings of the 16th Conference of the European Chapter
  of the Association for Computational Linguistics: Main Volume}, pages
  2190--2202. Association for Computational Linguistics.

\bibitem[{Xu et~al.(2021)Xu, Ebner, Yarmohammadi, White, Van~Durme, and
  Murray}]{xu2021gradual}
Haoran Xu, Seth Ebner, Mahsa Yarmohammadi, Aaron~Steven White, Benjamin
  Van~Durme, and Kenton Murray. 2021.
\newblock \href {https://aclanthology.org/2021.adaptnlp-1.22} {Gradual
  fine-tuning for low-resource domain adaptation}.
\newblock In \emph{Proceedings of the Second Workshop on Domain Adaptation for
  NLP}, pages 214--221. Association for Computational Linguistics.

\bibitem[{Yamakata et~al.(2020)Yamakata, Mori, and
  Carroll}]{yamakata2020english}
Yoko Yamakata, Shinsuke Mori, and John~A Carroll. 2020.
\newblock \href {https://aclanthology.org/2020.lrec-1.638} {English recipe flow
  graph corpus}.
\newblock In \emph{Proceedings of the 12th Language Resources and Evaluation
  Conference}, pages 5187--5194. European Language Resources Association.

\bibitem[{Zellers et~al.(2019)Zellers, Holtzman, Bisk, Farhadi, and
  Choi}]{zellers2019hellaswag}
Rowan Zellers, Ari Holtzman, Yonatan Bisk, Ali Farhadi, and Yejin Choi. 2019.
\newblock \href {https://doi.org/10.18653/v1/P19-1472} {{H}ella{S}wag: Can a
  machine really finish your sentence?}
\newblock In \emph{Proceedings of the 57th Annual Meeting of the Association
  for Computational Linguistics}, pages 4791--4800. Association for
  Computational Linguistics.

\bibitem[{Zhang et~al.(2020{\natexlab{a}})Zhang, Lyu, and
  Callison-Burch}]{zhang2020intent}
Li~Zhang, Qing Lyu, and Chris Callison-Burch. 2020{\natexlab{a}}.
\newblock \href {https://aclanthology.org/2020.aacl-main.35} {Intent detection
  with {W}iki{H}ow}.
\newblock In \emph{Proceedings of the 1st Conference of the Asia-Pacific
  Chapter of the Association for Computational Linguistics and the 10th
  International Joint Conference on Natural Language Processing}, pages
  328--333. Association for Computational Linguistics.

\bibitem[{Zhang et~al.(2020{\natexlab{b}})Zhang, Lyu, and
  Callison-Burch}]{zhang2020reasoning}
Li~Zhang, Qing Lyu, and Chris Callison-Burch. 2020{\natexlab{b}}.
\newblock \href {https://doi.org/10.18653/v1/2020.emnlp-main.374} {Reasoning
  about goals, steps, and temporal ordering with {W}iki{H}ow}.
\newblock In \emph{Proceedings of the 2020 Conference on Empirical Methods in
  Natural Language Processing}, pages 4630--4639. Association for Computational
  Linguistics.

\bibitem[{Zhou et~al.(2022)Zhou, Zhang, Yang, Lyu, Yin, Callison-Burch, and
  Neubig}]{zhou2022show}
Shuyan Zhou, Li~Zhang, Yue Yang, Qing Lyu, Pengcheng Yin, Chris Callison-Burch,
  and Graham Neubig. 2022.
\newblock \href {https://doi.org/10.18653/v1/2022.acl-long.214} {Show me more
  details: Discovering hierarchies of procedures from semi-structured web
  data}.
\newblock In \emph{Proceedings of the 60th Annual Meeting of the Association
  for Computational Linguistics (Volume 1: Long Papers)}, pages 2998--3012.
  Association for Computational Linguistics.

\bibitem[{Zhou et~al.(2019)Zhou, Shah, and Schockaert}]{zhou2019learning}
Yilun Zhou, Julie Shah, and Steven Schockaert. 2019.
\newblock \href {https://aclanthology.org/W19-5808} {Learning household task
  knowledge from {W}iki{H}ow descriptions}.
\newblock In \emph{Proceedings of the 5th Workshop on Semantic Deep Learning
  (SemDeep-5)}, pages 50--56. Association for Computational Linguistics.

\end{thebibliography}


\end{document}